\documentclass[11pt]{article}
\usepackage{emnlp2016}
\usepackage{times}
\usepackage{url}
\usepackage{latexsym}
\usepackage{multirow}
\usepackage{amssymb}

\usepackage{amsthm}
\usepackage{color}

\usepackage{amsmath}
\usepackage{verbatim}
\usepackage{tikz-qtree}
\usepackage{tikz}

\usepackage{filecontents}

\usetikzlibrary{shapes.geometric}
\usetikzlibrary{shapes.arrows}
\usetikzlibrary{patterns}
\usetikzlibrary{chains}
\usetikzlibrary{calc}

\emnlpfinalcopy

\newcommand{\ignore}[1]{}

\newcommand{\nascomment}[1]{\textcolor{blue}{\textbf{[#1 --\textsc{nas}]}}}

\newsavebox{\one}
\newsavebox{\two}
\newsavebox{\three}
\newsavebox{\four}
\newsavebox{\five}

\ifx\release\undefined
    
\fi

\begin{document}


%

\title{Training with Exploration Improves a Greedy Stack LSTM Parser}
\author{Miguel Ballesteros$^{\diamondsuit}$ ~ Yoav Goldberg$^{\clubsuit}$~ Chris Dyer$^{\spadesuit}$ ~ Noah A. Smith$^{\heartsuit}$\\
$^\diamondsuit$NLP Group, Pompeu Fabra University, Barcelona, Spain \\
$^\clubsuit$Computer Science Department, Bar-Ilan University, Ramat Gan, Israel \\
$^\spadesuit$Google DeepMind, London, UK \\
$^{\heartsuit}$Computer Science \& Engineering, University of Washington, Seattle, WA, USA\\
{ \sf miguel.ballesteros@upf.edu, yoav.goldberg@gmail.com, } \\
{\sf  cdyer@google.com, nasmith@cs.washington.edu}
}

\maketitle              

\begin{abstract}
    We adapt the greedy stack LSTM dependency parser of \newcite{lstmacl15} to support a
    training-with-exploration procedure using dynamic oracles
    \cite{goldberg2013training} instead of assuming an error-free
    action history. This form of training, which accounts for model predictions at training time, improves
    parsing accuracies.  We discuss some modifications needed in order
    to get training with exploration to work well for a probabilistic
    neural network
    dependency parser.
\end{abstract}

\section{Introduction}

Natural language parsing can be formulated as a series of decisions that read
words in sequence and incrementally combine them to form syntactic structures;
this formalization is known as transition-based parsing, and is often coupled
with a greedy search procedure
\cite{yamada03,nivre03iwpt,nivre2004,nivre08cl}. 
The literature on transition-based parsing is vast, but all works share in
common a classification component that takes into account features of the current
parser state\footnote{The term ``state''
 refers to the collection of previous decisions (sometimes called the history),
 resulting partial structures, which are typically stored in a stack data structure,
 and the words remaining to be processed.} and predicts the next action to
 take conditioned on the state. The state is of unbounded size.

Dyer et al.~\shortcite{lstmacl15} presented a parser in which the
parser's unbounded state is embedded in a fixed-dimensional continuous space using recurrent neural networks. Coupled with a recursive
tree composition function, the feature representation is able to capture
information from the entirety of the state, without resorting to locality
assumptions that were common in most other transition-based parsers.
The use of a novel stack LSTM
data structure allows the parser to maintain a constant time per-state update,
and retain an overall linear parsing time. 

The Dyer et al.~parser was trained to maximize the likelihood of
gold-standard transition sequences, given words. At test time, the
parser makes \emph{greedy} decisions according to the learned
model. Although this setup obtains very good performance, the training
and testing conditions are mismatched in the following way: at
training time the historical context of an action is always derived
from the gold standard (i.e., perfectly correct past actions), but at test time, it will be a model prediction.

In this work, we adapt the training criterion so as to explore parser states drawn not only from the training data, but also from the model as it is being learned.
To do so, we use the method of Goldberg and Nivre \shortcite{goldberg12dynamic,goldberg2013training} to dynamically chose an optimal (relative to the final attachment accuracy) action given an imperfect history. By interpolating between algorithm states sampled from the model and those sampled from the training data, more robust predictions at test time can be made. We show that
the technique can be used to improve the strong parser of Dyer
et al.

\ignore{
\paragraph{Notation} We follow the convention that vectors are written with lowercase, boldface letters (e.g., $\mathbf{v}$ or $\mathbf{v}_w$); matrices are written with uppercase, boldface letters (e.g., $\mathbf{M}$, $\mathbf{M}_{a}$, or $\mathbf{M}_{ab}$), and scalars are written as lowercase letters (e.g., $s$ or $q_z$). Structured objects such as sequences of discrete symbols are written with lowercase, bold, italic letters (e.g., $\boldsymbol{w}$ refers to a sequence of input words). 
We use a semicolon to denote vector concatenation (e.g., $[\mathbf{a};\mathbf{b}]$).
}

\section{Parsing Model and Parameter Learning}
Our departure point is the parsing model described by
\newcite{lstmacl15}.
We do not describe the model in detail, and refer the reader to the original
work.
At each stage $t$ of the parsing process, the parser state is encoded into a vector
$\mathbf{p}_t$, which is used to compute the probability of the parser action at
time $t$ as:
\begin{align}
p(z_t \mid \mathbf{p}_t) = \frac{\exp \left( \mathbf{g}_{z_t}^{\top}
\mathbf{p}_t + q_{z_t} \right)}{\sum_{z' \in \mathcal{A}(S,B)} \exp \left(
\mathbf{g}_{z'}^{\top} \mathbf{p}_t + q_{z'} \right)},
\label{eq:local-objective}
\end{align}
where $\mathbf{g}_z$ is a column vector representing the (output) embedding of
the parser action $z$, and $q_z$ is a bias term for action $z$. The set
$\mathcal{A}(S,B)$ represents the valid transition actions that may be taken in
the current state.
Since $\mathbf{p}_t$ encodes information about all
previous decisions made by the parser, the chain rule gives
the probability of any valid sequence of parse transitions $\boldsymbol{z}$ conditional on the input:
\begin{align}
p(\boldsymbol{z} \mid \boldsymbol{w}) = \prod_{t=1}^{|\boldsymbol{z}|} p(z_t \mid \mathbf{p}_t). \label{eq:objective}
\end{align}


The parser is trained to maximize the conditional probability of taking a
``correct'' action at each parsing state.  The definition of what
constitutes a ``correct'' action is the major difference between a static oracle
as used by \newcite{lstmacl15} and the dynamic oracle explored here.

Regardless of the oracle, our training implementation constructs a
computation graph (nodes that represent values, linked by directed
edges from each function's inputs to its outputs) for the negative log
probability for the oracle transition sequence as a function of the current
model parameters and uses forward- and
backpropagation to obtain the gradients respect to the model parameters
\cite[section 4]{lecun:1998}. 

\subsection{Training with Static Oracles} \label{static-oracle}
With a static oracle, the training procedure computes a canonical
reference series of transitions for each gold parse tree. It then runs the parser through
this canonical sequence of transitions, while keeping track of the state
representation $\mathbf{p}_t$ at each step $t$, as well as the distribution over transitions $p(z_t
\mid \mathbf{p}_t)$ which is predicted by the current classifier for the state representation.
Once the end of the sentence is reached, the parameters are  updated towards
maximizing the likelihood of the reference transition sequence
(Equation~\ref{eq:objective}), which equates to maximizing the
probability of the correct transition,
$p(z_{g_t} \mid \mathbf{p_t})$, at each state along the path.

\subsection{Training with Dynamic Oracles}
\label{dyn-train}
In the static oracle case, the parser is trained to predict the best transition
to take at each parsing step,
assuming all previous transitions were correct.
Since the parser is likely to make mistakes at test time and encounter states
it has not seen during training, this training criterion is problematic \cite[\emph{inter alia}]{daume09,ross11,goldberg12dynamic,goldberg2013training}. Instead, we would prefer to train the parser to behave
optimally even after making a mistake (under the constraint that it
cannot backtrack or fix any previous decision).  We thus need to include in the
training examples states that result from wrong parsing decisions,
together with the optimal transitions to take in these states.
To this end we reconsider which training examples to show, and what it
means to behave optimally on these training examples.
The framework of training with exploration using dynamic oracles suggested by
Goldberg and Nivre \shortcite{goldberg12dynamic,goldberg2013training} provides
answers to these questions. While the application of dynamic oracle training is
relatively straightforward, some adaptations were needed to accommodate the
probabilistic training objective.  These adaptations mostly follow
Goldberg \shortcite{goldberg2013calibrated}.

\paragraph{Dynamic Oracles.} A \emph{dynamic oracle} is the component that, given a gold parse tree, 
provides the optimal set of possible actions to take
for any valid parser state.  In contrast to static oracles that derive a
canonical state sequence for each gold parse tree and say nothing about
states that deviate from this canonical path, the dynamic
oracle is well defined for states that result from parsing
mistakes, and they may produce more than a single gold action for a given
state.  Under the dynamic oracle framework, an action is said to be
optimal for a state if the best tree that can be reached after taking the
action is no worse (in terms of accuracy with respect to the gold tree) than the
best tree that could be reached prior to taking that action.

Goldberg and Nivre \shortcite{goldberg2013training} define the arc-decomposition
property of transition systems, and show how to derive efficient
dynamic oracles for transition systems that are arc-decomposable.\footnote{Specifically: 
for every parser configuration $\mathbf{p}$ and group of arcs $A$, if
each arc in $A$ can be derived from $\mathbf{p}$, then a valid tree structure
containing \emph{all} of the arcs in $A$ can also be derived from $\mathbf{p}$.
This is a sufficient condition, but whether it is necessary is
unknown; hence the question of an efficient, $O(1)$ dynamic oracle for the augmented system is open. \label{fn:arc-decomposition}}
Unfortunately,
the arc-standard transition system does not have this property.  While it is
possible to compute dynamic oracles for the arc-standard system
\cite{goldberg2013tabular}, the computation relies on a dynamic programming
algorithm which is polynomial in the length of the stack.
As the dynamic oracle has to be queried for each parser state seen during
training, the use of this dynamic oracle will make the training runtime
several times longer.
We chose instead to switch to the arc-hybrid transition system \cite{kuhlmann11dynamic}, 
which is very
similar to the arc-standard system but is arc-decomposable and hence admits an
efficient $O(1)$ dynamic oracle, resulting in only negligible increase
to training runtime.
We implemented the dynamic oracle to the arc-hybrid system as described by
Goldberg \shortcite{goldberg2013training}.

\paragraph{Training with Exploration.} In order to expose the parser to configurations that are likely to result from
incorrect parsing decisions, we make use of the probabilistic nature of the
classifier.
During training, instead of following the gold action, 
we sample the next transition according to
the output distribution the classifier assigns to the current
configuration. Another option, taken by Goldberg and Nivre, is to
follow the one-best action predicted by the classifier.  However, initial
experiments showed that the one-best approach did not work well. Because the
neural network classifier becomes accurate early on in the training process, the
one-best action is likely to be correct, and the parser is then exposed to very
few error states in its training process. By sampling from the predicted
distribution, we are effectively increasing the chance of straying from the gold
path during training, while still focusing on mistakes that receive relatively
high parser scores.
We believe further formal analysis of this method will reveal
connections to reinforcement learning and, perhaps, other methods for
learning complex policies.

Taking this idea further, we could increase the number of error-states observed
in the training process by changing the sampling distribution so
as to bias it toward more low-probability states. We do this by raising each
probability to the power of $\alpha$ ($0 < \alpha \leq 1$) and
re-normalizing. 
This transformation keeps the relative ordering of the events, while shifting
probability mass towards less frequent events.  As we show below, this turns out
to be very beneficial for the configurations that make use of external embeddings.
Indeed, these configurations achieve high accuracies and sharp class distributions 
early on in the training process.

The parser is trained to maximize the likelihood of a correct action $z_g$
at each parsing state $\mathbf{p}_t$ according to Equation~\ref{eq:local-objective}.  When using the dynamic oracle, a state $\mathbf{p}_t$
may admit multiple correct actions $\boldsymbol{z_g} = \{z_{g_i}, \ldots, z_{g_k}\}$. Our
objective in such cases is the marginal likelihood of all correct actions,\footnote{A similar objective was used by Riezler et
al~\shortcite{riezler2000lexicalized},
Charniak and Johnson~\shortcite{charniak05} and Goldberg~\shortcite{goldberg2013calibrated} in the
context of log-linear probabilistic models.
}
\begin{align}
    p(\boldsymbol{z_g} \mid \mathbf{p}_t) = \sum_{z_{g_i} \in \boldsymbol{z_g}}
    p(z_{g_i} \mid \mathbf{p}_t) .
\end{align}
\ignore{\nascomment{small notational change here to boldsymbol rather than
  mathbf for consistency with what we said in the ``Notation'' section}}

\section{Experiments}

Following the same settings of Chen and Manning \shortcite{chen:2014} and Dyer et al \shortcite{lstmacl15} we report results\footnote{The results on the development sets are similar and only used for optimization and validation.} in the English PTB
and Chinese CTB-5. 
Table \ref{dyn-sd} shows the results of the parser in its different configurations. The
table also shows the best result obtained with the static oracle (obtained by rerunning Dyer et al. parser) for the sake of comparison between static and dynamic training strategies.

\begin{table}[!ht]
\begin{center}
\scalebox{0.75}{
\begin{tabular}{|l|cc|cc|}
 \multicolumn{1}{l}{} & \multicolumn{2}{c}{English} & \multicolumn{2}{c}{Chinese}\\
\cline{2-5}
\multicolumn{1}{l|}{Method} & UAS & LAS & UAS & LAS\\
\hline
Arc-standard  (Dyer et al.) & 92.40 & 90.04 & 85.48 & 83.94 \\
Arc-hybrid (static) & 92.08 & 89.80 & 85.66 & 84.03 \\
Arc-hybrid (dynamic) & 92.66 & 90.43 & 86.07 & 84.46 \\
Arc-hybrid (dyn., $\alpha = 0.75$) & 92.73 & 90.60 & 86.13 & 84.53 \\
\hline                    
\multicolumn{5}{l}{+ pre-training:}\\
\hline
Arc-standard  (Dyer et al.) & 93.04 & 90.87 & 86.85 & 85.36 \\
Arc-hybrid (static)  & 92.78 & 90.67 & 86.94 & 85.46 \\
Arc-hybrid (dynamic)  & 93.15 & 91.05 & 87.05 & 85.63 \\
Arc-hybrid (dyn., $\alpha = 0.75$)& \bf93.56 & \bf91.42 & \bf87.65 & \bf86.21 \\
\hline

\end{tabular} \hspace{0.25in}
}\end{center}

\caption{Dependency parsing: English (SD) and Chinese.  
\label{dyn-sd}}

\end{table}

\begin{table*}[!ht]
\begin{center}
\scalebox{0.72}{
\begin{tabular}{|l|cc|cc|cc|cc|cc|cc|cc|}
 \multicolumn{1}{c}{} & \multicolumn{2}{c}{Catalan} & \multicolumn{2}{c}{Chinese} & \multicolumn{2}{c}{Czech} & \multicolumn{2}{c}{English} & \multicolumn{2}{c}{German} & \multicolumn{2}{c}{Japanese} & \multicolumn{2}{c}{Spanish} \\
\cline{2-15}
\multicolumn{1}{l|}{Method} & UAS & LAS & UAS & LAS & UAS & LAS & UAS & LAS & UAS & LAS & UAS & LAS & UAS & LAS \\
\hline

Arc-standard, static + PP  & 89.60 & 85.45 & 79.68 & 75.08 & 77.96 & 71.06 & 91.12 & 88.69 & 88.09 & 85.24 & 93.10 & 92.28 & 89.08 & 85.03\\
+ pre-training & -- & -- & 82.45 & 78.55 & -- & -- & 91.59 & 89.15 & 88.56 & 86.15 & -- & -- & 90.76 & 87.48 \\
Arc-hybrid, dyn.~+ PP & 90.45 & 86.38 & 80.74 & 76.52 & 85.68 & 79.38 & 91.62 & 89.23 & 89.80 & 87.29 & 93.47 & 92.70 & 89.53 & 85.69 \\
+ pre-training & -- & -- & \bf83.54 & \bf79.66 & -- & -- & \bf92.22 & \bf89.87 & \bf90.34 & \bf88.17 & -- & -- & \bf91.09 & 87.95 \\
\hline                    
Y'15  & -- & -- & -- & -- & 85.2 & 77.5 & 90.75 & 88.14 & 89.6 & 86.0 & -- & -- & 88.3 & 85.4\\
A'16 + pre-training & \bf91.24 & \bf88.21 & 81.29 & 77.29 & \bf85.78 & \bf80.63 & 91.44 & 89.29 & 89.12 & 86.95 & \bf93.71 & \bf92.85 & 91.01 & \bf88.14\\
\hline                   
A'16-beam & 92.67 & 89.83 & 84.72 & 80.85 & 88.94 & 84.56 & 93.22 & 91.23 & 90.91 & 89.15 & 93.65 & 92.84 & 92.62 & 89.95\\
\hline
\end{tabular} 
}
\end{center}
\caption{Dependency parsing results.  The dynamic oracle uses $\alpha
  = 0.75$ (selected on English; see Table~\ref{dyn-sd}). PP refers to pseudo-projective
  parsing. Y'15 and A'16  are beam = 1 parsers from
  Yazdani and Henderson (2015) and
  Andor et al. (2016), respectively. A'16-beam is the parser with beam larger than 1 by
  Andor et al. (2016). Bold numbers indicate the best results among the greedy parsers. \label{table-res-2}}
\end{table*}

The score achieved by the dynamic oracle for English is 93.56 UAS. This is remarkable given that
the parser uses a completely greedy search procedure. Moreover, the
Chinese score establishes the state-of-the-art, using the same settings as \newcite{chen:2014}.  

The error-exploring dynamic-oracle training always improves over static oracle training controlling for the transition system, but the arc-hybrid system slightly under-performs
the arc-standard system when trained with static oracle. Flattening the
sampling distribution ($\alpha=0.75$) is
especially beneficial when training with pretrained word embeddings.


In order to be able to compare with similar greedy parsers
\cite{yazdani-henderson:2015:CoNLL,andor-16}\footnote{We report the
  performance of these parsers in the most comparable setup, that is,
  with beam size 1 or greedy search.} we report the performance of
the parser on the multilingual treebanks of the CoNLL 2009 shared task
\cite{conll2009}. Since some of the treebanks contain nonprojective
sentences and arc-hybrid does not allow nonprojective trees, we use
the pseudo-projective approach \cite{nivre05acl}. We used predicted
part-of-speech tags provided by the CoNLL 2009 shared task
organizers. We also include results with pretrained word embeddings
for English, Chinese, German, and Spanish following the same training
setup as Dyer et al.~(2015); for English and Chinese we used the same
pretrained word embeddings as in Table~\ref{dyn-sd}, for German we
used the monolingual training data from the WMT 2015
dataset\ignore{\footnote{\url{http://www.statmt.org/wmt15/translation-task.html}}}
and for Spanish we used the Spanish Gigaword version~3. See Table~\ref{table-res-2}.

\section{Related Work}
\label{sec:related}

Training greedy parsers on non-gold outcomes, facilitated by 
dynamic oracles, has been explored by several researchers in different ways
\cite{goldberg12dynamic,goldberg2013training,goldberg2013tabular,honnibal2013non,honnibal2014joint,gomez2014polynomial,bjorkelund2015non,tokgoz2015transition,gomez2015efficient,TACL659}.
More generally, training greedy search systems by paying attention to the
expected classifier behavior during test time has been explored under the
imitation learning and learning-to-search frameworks
\cite{abbeel2004apprenticeship,daume05,vlachos12,he12,daume09,ross11,icml2015_changb15}. Directly modeling the probability of making a mistake has also been explored for parsing~\cite{yazdani-henderson:2015:CoNLL}. Generally, the use of RNNs to conditionally predict actions in
sequence given a history is spurring increased interest in training
regimens that make the learned model more robust to test-time
prediction errors. Solutions based on curriculum learning
\cite{bengio2015scheduled}, expected loss training \cite{shen:2015},
and reinforcement learning have been proposed \cite{ranzato}. Finally,
abandoning greedy search in favor of approximate global search offers
an alternative solution to the problems with greedy
search~\cite{andor-16}, and has been analyzed as well
\cite{Kulesza2007,Finley2008}, including for parsing \cite{martins-09}.

\ignore{Time and test time prediction was recently explored by \newcite{bengio2015scheduled} for the task of language modeling.
However, to the best of
our knowledge this is the first time that such a training procedure is explored
together with neural networks for parsing. The results show that the consistent
gains in accuracy due to dynamic oracle training that were observed in previous work
persist also for the very strong LSTM-based parser,
although some care had to be taken to adapt the training procedures to the
probabilistic nature of the objective, and to the high accuracy levels of the
underlying model.
}

\section{Conclusions}

\newcite{lstmacl15} presented stack LSTMs and used them to implement a
transition-based dependency parser. The parser uses a greedy learning
strategy which potentially provides very high parsing speed while
still achieving state-of-the-art results. We have demonstrated that
improvement by training the greedy parser on
non-gold outcomes; dynamic oracles improve the stack LSTM parser, achieving 93.56 UAS for English, maintaining greedy search.

\section*{Acknowledgments}
 This work was sponsored in part by the U.~S.~Army Research Laboratory and the
 U.~S.~Army Research Office under contract/grant number W911NF-10-1-0533, and in
 part by NSF CAREER grant IIS-1054319. Miguel Ballesteros was supported by the
 European Commission under the contract numbers FP7-ICT-610411 (project
 MULTISENSOR) and H2020-RIA-645012 (project KRISTINA). 
 Yoav Goldberg is supported by the Intel Collaborative Research Institute for
 Computational Intelligence (ICRI-CI), a Google Research Award and the Israeli Science Foundation
 (grant number 1555/15).

\bibliographystyle{emnlp2016}
\bibliography{biblio,main,dynamic}%
\end{document}